\documentclass[sigconf,nonacm]{acmart}
\pdfoutput=1

\usepackage[T1]{fontenc}
\usepackage[utf8]{inputenc}
\usepackage[english]{babel}

\usepackage{comment}
\usepackage{xcolor}
\AtBeginDocument{%
  \providecommand\BibTeX{{%
    \normalfont B\kern-0.5em{\scshape i\kern-0.25em b}\kern-0.8em\TeX}}}

\begin{document}

\title{Determining Accessible Sidewalk Width by Extracting Obstacle Information from Point Clouds}


\author{Cl{\'a}udia Fonseca Pinh{\~a}o, Chris Eijgenstein, Iva Gornishka, \\Shayla Jansen, Diederik M.\ Roijers, Daan Bloembergen}
\authornote{This project was a team effort, all authors contributed equally and the ordering of names is arbitrary.}
\email{{c.fonsecapinhao,c.eijgenstein,i.gornishka,shayla.jansen,d.roijers,d.bloembergen}@amsterdam.nl}
\affiliation{%
  \institution{Urban Innovation and R\&D, City of Amsterdam}
  \city{Amsterdam}
  \country{The Netherlands}
}

\patchcmd{\maketitle}{\@copyrightspace}{}{}{}
\maketitle

\section{Introduction}

Obstacles on the sidewalk,  such as benches, garbage bins, signs, vegetation, bicycles, and terraces, among others, often block the path, limiting passage and resulting in frustration and wasted time, especially for citizens and visitors who use assistive devices (wheelchairs, walkers, strollers, canes, etc.).  To enable equal participation and use of the city, all citizens should be able to perform and complete their daily activities {\color{black} without an unreasonable amount of effort}. Therefore, we aim to offer accessibility information regarding sidewalks, so that citizens can better plan their routes, and to help city officials identify the location of bottlenecks and act on them. 

In this paper we propose a novel pipeline to estimate obstacle-free sidewalk widths based on 3D point cloud data of the city of Amsterdam, as the first step to offer a more complete set of information regarding sidewalk accessibility.\footnote{Our code can be found at\\\url{https://github.com/Amsterdam-AI-Team/Urban_PointCloud_Sidewalk_Width}} This solution is especially relevant since the existing registers can be outdated or incomplete and require regular manual inspections to keep up with the constant changes in the city. In addition, obstacles may be privately owned and thus not part of city asset registries. Point clouds enable us to extract precise locations and measurements of objects, and have proven valuable for extracting and classifying objects in urban areas \cite{balado2018automatic,zhang2019review,guo2020deep}. Previous work has also used point clouds to measure e.g. sidewalk slope and width \cite{hou2020network}, however without taking obstacles into account. In contrast to e.g. specially developed sidewalk scanning tools \cite{frackelton2013measuring}, our approach is easily scalable for the entire city.

Our main contributions are as follows. First, we extract static obstacles on sidewalks by applying a change detection algorithm on two separate point cloud scans. We then compute paths along the sidewalk that take into account these extracted obstacles and record their minimal width. We compare this to the full sidewalk width, without obstacles. This analysis allows us to identify large discrepancies between the full widths and the effective widths of sidewalks in Amsterdam, and we would thus argue that this information is a key input for policymakers and users of the sidewalks alike.

\section{Definitions}

\begin{figure}[tb]
  \centering
  \includegraphics[width=\linewidth]{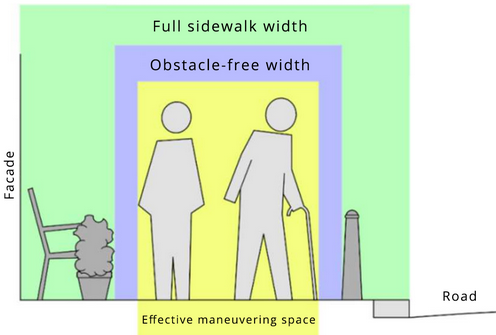}
  \caption[]{Image with full sidewalk width in green and obstacle-free width in purple (from Leidraad Centrale Verkeerscommissie\protect\footnotemark, in Dutch).}
  \label{fig:sidewalk}
\end{figure}

We first define three key concepts for accessible mobility on sidewalks, which are illustrated in Figure \ref{fig:sidewalk}. First, the effective manoeuvring space (yellow) is the minimal width that would fit the person or people moving over the sidewalk. The effective manoeuvring space needs to be smaller than the obstacle-free width (purple) of the sidewalk. Finally, the full sidewalk width (green) is the width of the sidewalk without taking obstacles into account.  

The minimal policy norms for sidewalks are: a minimum width of $0.9m$ required for users of mobility devices; at least $1.8m$ to allow $10$ pedestrians per minute or for two mobility device users to comfortably pass each other; and at least $2.9m$ for sidewalks that have $30$ pedestrians per minute.

\footnotetext{\url{https://bikecity.amsterdam.nl/documents/52/Leidraad_CVC_2020.pdf}, p. 17.}

\begin{figure}[tb]
  \centering
  \includegraphics[width=\linewidth]{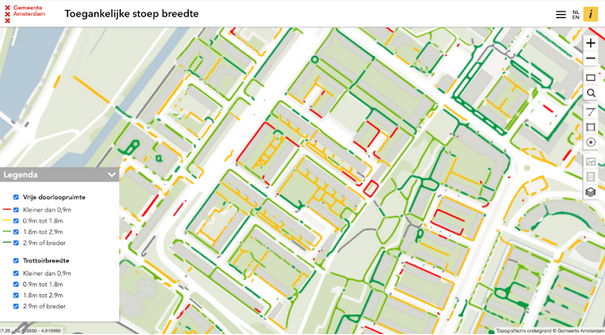}
  \caption{Mock-up visualisation of obstacle-free widths in 4 colours: red for $<0.9m$, yellow for $0.9-1.8m$, light green for $1.8-2.9m$, and dark green for $>2.9m$.}
  \label{fig:vis}
\end{figure}

In this paper, we estimate both the full sidewalk width and the obstacle-free width, and compare them to the above-mentioned norms. These results can be processed into a map (Figure \ref{fig:vis}) that contains information about the (possibly limited) accessibility of the sidewalks.  

\section{Data sources and processing}
\label{sec:data}

To estimate the full sidewalk widths and obstacle-free widths we employ different data sources. Firstly, we use the municipal database on topographical data to compute the full sidewalk width. Secondly, we use point cloud data resulting from LIDAR scans to detect obstacles, in order to estimate the obstacle-free width. 

\paragraph{Topographical data} The primary preexisting data we use is the municipal basic registration on large-scale topographical data (in Dutch: basisregistratie grootschalige topografie or BGT).\footnote{\url{https://www.amsterdam.nl/bestuur-organisatie/organisatie/dienstverlening/basisinformatie/basisinformatie/registraties/bgt/}} The BGT contains the shapes (polygons) of all sidewalks in the city.\footnote{\color{black} If such data is not available, it may be possible to detect the shapes of sidewalks automatically, e.g., using satellite data \cite{hosseini2021sidewalk}.}  The BGT further contains information regarding known static objects, such as trees and terraces. Objects larger than $10m^2$, such as flower beds are not counted as sidewalk. We disregard smaller objects in this phase of the pipeline as we regard the point cloud data as the most accurate and up-to-date.

\paragraph{Point cloud data} To calculate the obstacle-free widths, we need accurate and up-to-date data on all possible obstacles. For many such obstacles, we do not have an extensive record in municipal databanks, including the exact current shape and location of these objects. Furthermore, many obstacles are owned by private citizens or businesses and thus are not included in the municipal registries at all. In order to find and measure all these obstacles, we use street-level 3D point cloud data provided by CycloMedia\footnote{\url{http://www.cyclomedia.com/}}, which was generated using a LIDAR scanner mounted on top of a car. We use two separate scans of the city, both collected in the summer of 2020. {\color{black} A key advantage of point cloud data over, e.g., street level images, is that the points can be easily mapped to exact geo-locations. Furthermore, compared to satellite data, it suffers less from occlusions by e.g., overhanging trees.}

\section{Obstacle Extraction}
\label{sec:extract}
\begin{figure}[tb]
  \centering
  \includegraphics[width=\linewidth]{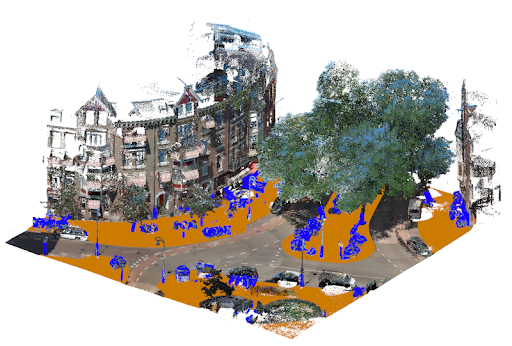}
  \caption{Example of obstacles (blue) above the public sidewalk (brown) in a point cloud.}
  \label{fig:pointcloud}
\end{figure}

First, all points in the point cloud above the sidewalk are extracted, between the ground up to the height that could be an obstacle for a pedestrian, i.e., $2m$. This range covers both floor-level objects and overhanging obstacles such as bushes and trees, as illustrated in Figure \ref{fig:pointcloud}.
To filter out the points that are part of the sidewalk itself (ground, pavement), we use AHN4\footnote{\url{https://www.ahn.nl/}} elevation data to automatically label ground surface points \cite{bloembergen2021automatic}. 

It is key to include only static obstacles - such as benches and trees - while ignoring temporary obstacles - such as people or incidentally parked bicycles. To achieve this, we use two separate scans, recorded with a roughly three-month interval, and apply the M3C2 change detection algorithm \cite{lague2013accurate}. This algorithm compares two point clouds spatially and identifies points that have changed from one scan to the next. Only the obstacles that are visible in both scans are kept in the final data set. However, it can still happen that temporary obstacles exist in the same location in both scans, such as parked bicycles or construction sites, and thus are incorrectly seen as static obstacles.

\begin{figure}[tb]
  \centering
  \includegraphics[width=0.6\linewidth]{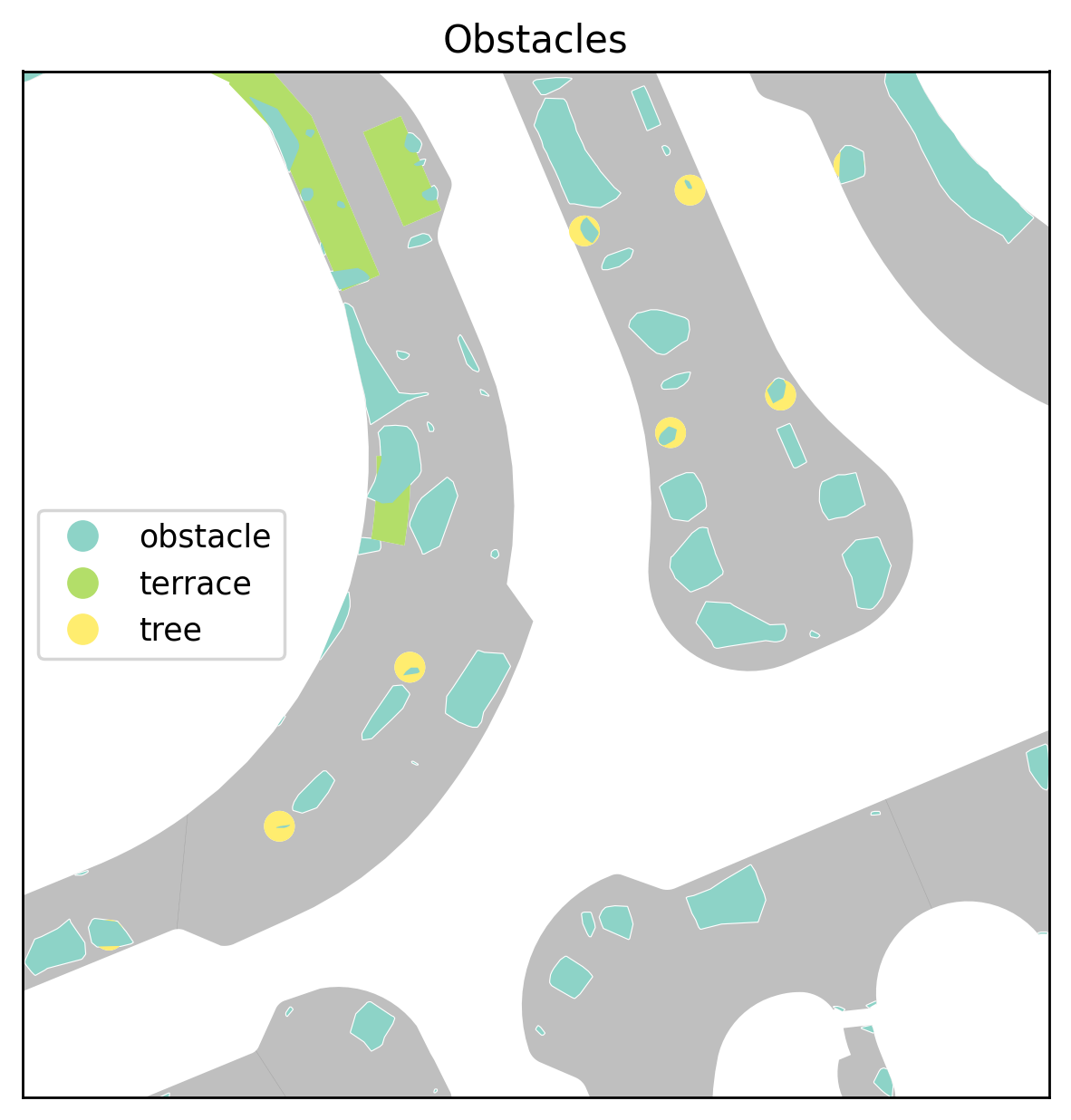}
  \caption{2D polygons generated from the obstacles in the point cloud (blue), along with terraces (green) and trees (yellow) taken from municipal data sources.}
  \label{fig:poly}
\end{figure}
\begin{figure}[tb]
  \centering
  \includegraphics[width=\linewidth]{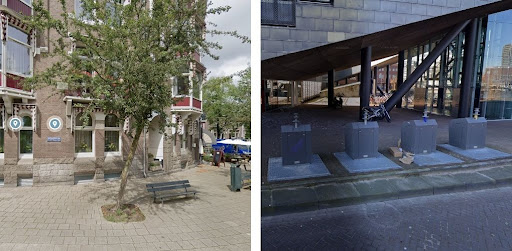}
  \caption{Example of a tree bed and container bases (from Google Maps).}
  \label{fig:treebin}
\end{figure}

We generate a 2D projection of all static obstacles we find in this way, by computing a closely fitting polygon around each cluster of points that represent an obstacle (Figure \ref{fig:poly}). In addition, we add known static obstacles for which we have accurate location data, such as trees and underground container bases. During validation field trips, we noticed their sizes were often underestimated by our method, since the tree-beds and container bases are almost level with the ground and thus seen as ground surface by our method, while they do have a potentially significant impact on sidewalk accessibility (Figure \ref{fig:treebin}).

Finally, we also added the shapes of terraces of bars and restaurants from the municipality permit data, as these often have tables and chairs at different places at different times and are therefore hard to detect robustly using our change detection algorithm.

\section{Width calculation}
The next step is to use the sidewalk shapes with and without obstacles to compute the obstacle-free width and full width of all sidewalks. We base our width calculation procedure on the Sidewalk Widths NYC project (SWNYCP) \cite{sidewalkGH}. It uses the Centerline python package to create paths following the centre of each sidewalk polygon. Some adjustments are made to these paths, like simplifying them and removing short dead-ends. Furthermore, centerlines are divided up into segments if they are longer than $10m$, to avoid generalising over too-long paths while maintaining a moderate level of detail on the output maps. Finally, it calculates the minimum width encountered on each of the resulting segments. 

\begin{figure}[tb]
  \centering
  \setlength\tabcolsep{1pt}
  \begin{tabular}{cc}
    \includegraphics[width=0.49\linewidth]{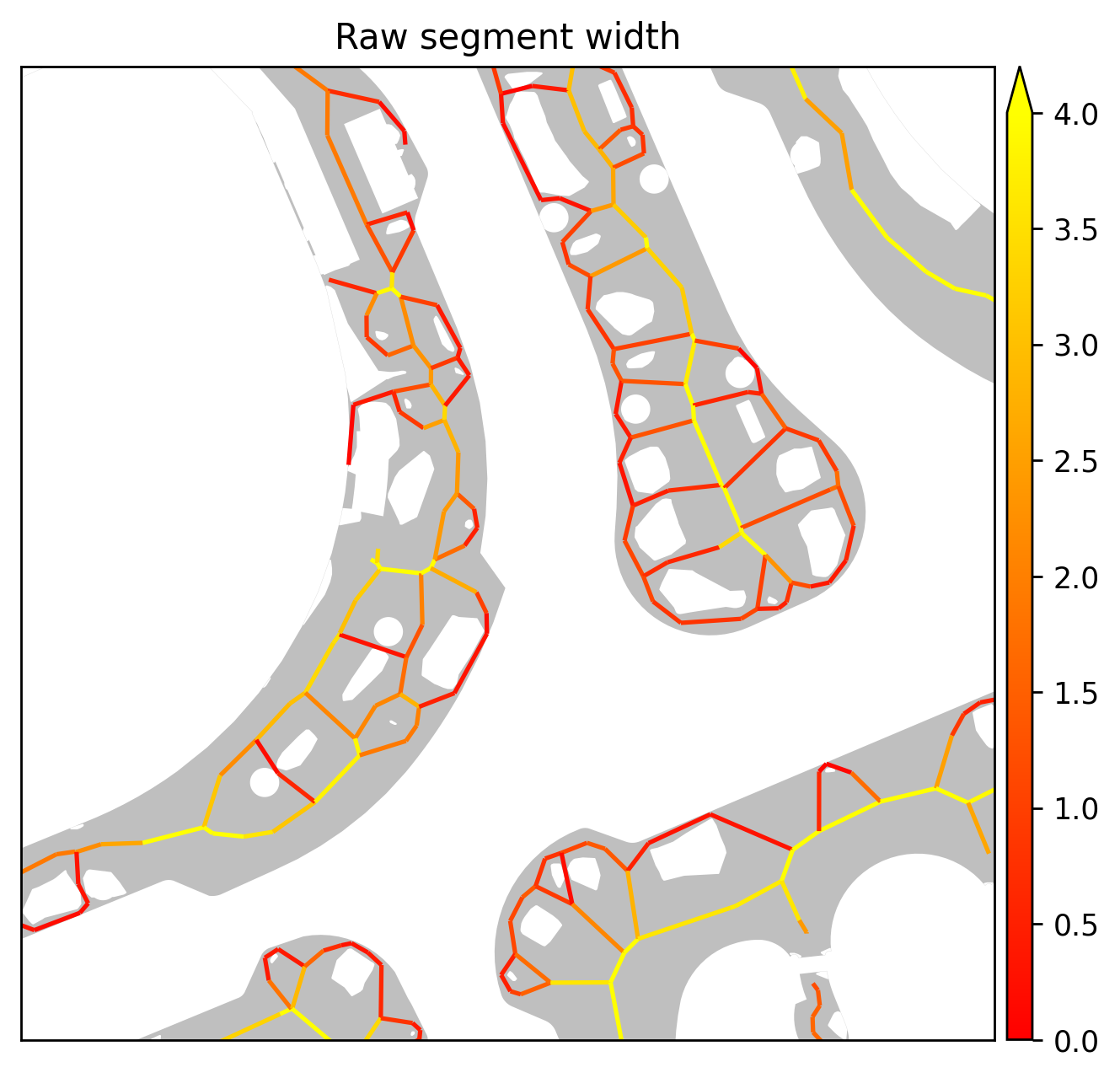} & \includegraphics[width=0.49\linewidth]{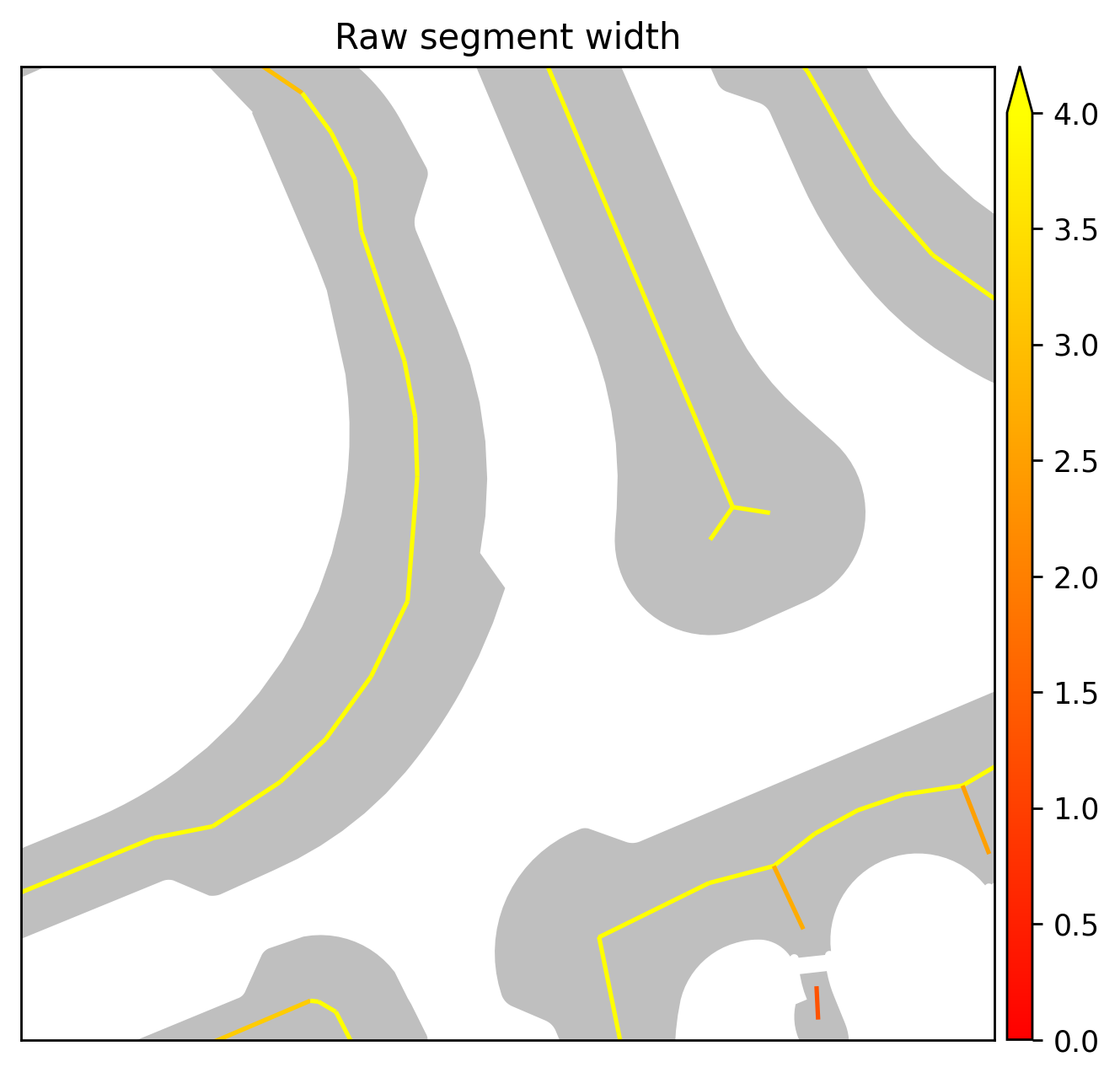} \\
  \end{tabular}
  \caption{Example of the calculated paths and their widths (in meters) with and without obstacles.}
  \label{fig:pathpath}
\end{figure}

Contrary to SWNYCP, we aim to calculate and compare both the full sidewalk widths and the obstacle-free widths. Therefore we perform the width calculation with and without obstacles. For the sidewalks with obstacles, the resulting set of paths contains short detailed paths around all obstacles (Figure \ref{fig:pathpath} left). On the other hand, if we apply the same procedure to the sidewalks without obstacles, we get much coarser paths (Figure \ref{fig:pathpath} right). We refer to these coarser paths as the \emph{major paths} as this level of detail is most suitable to show to the users.

\begin{figure}[tb]
  \centering
  \includegraphics[width=\linewidth]{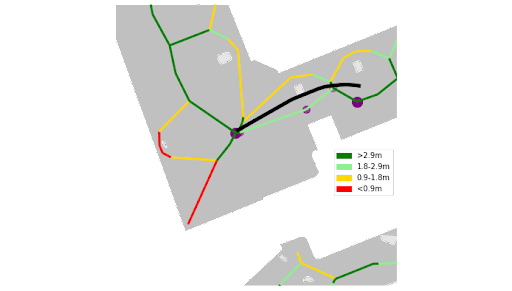}
  \caption{Illustration of the mapping of detailed paths around obstacles to major paths.}
  \label{fig:2major}
\end{figure}

\begin{figure}[tb]
  \centering
  \setlength\tabcolsep{1pt}
  \begin{tabular}{cc}
    \includegraphics[width=0.49\linewidth]{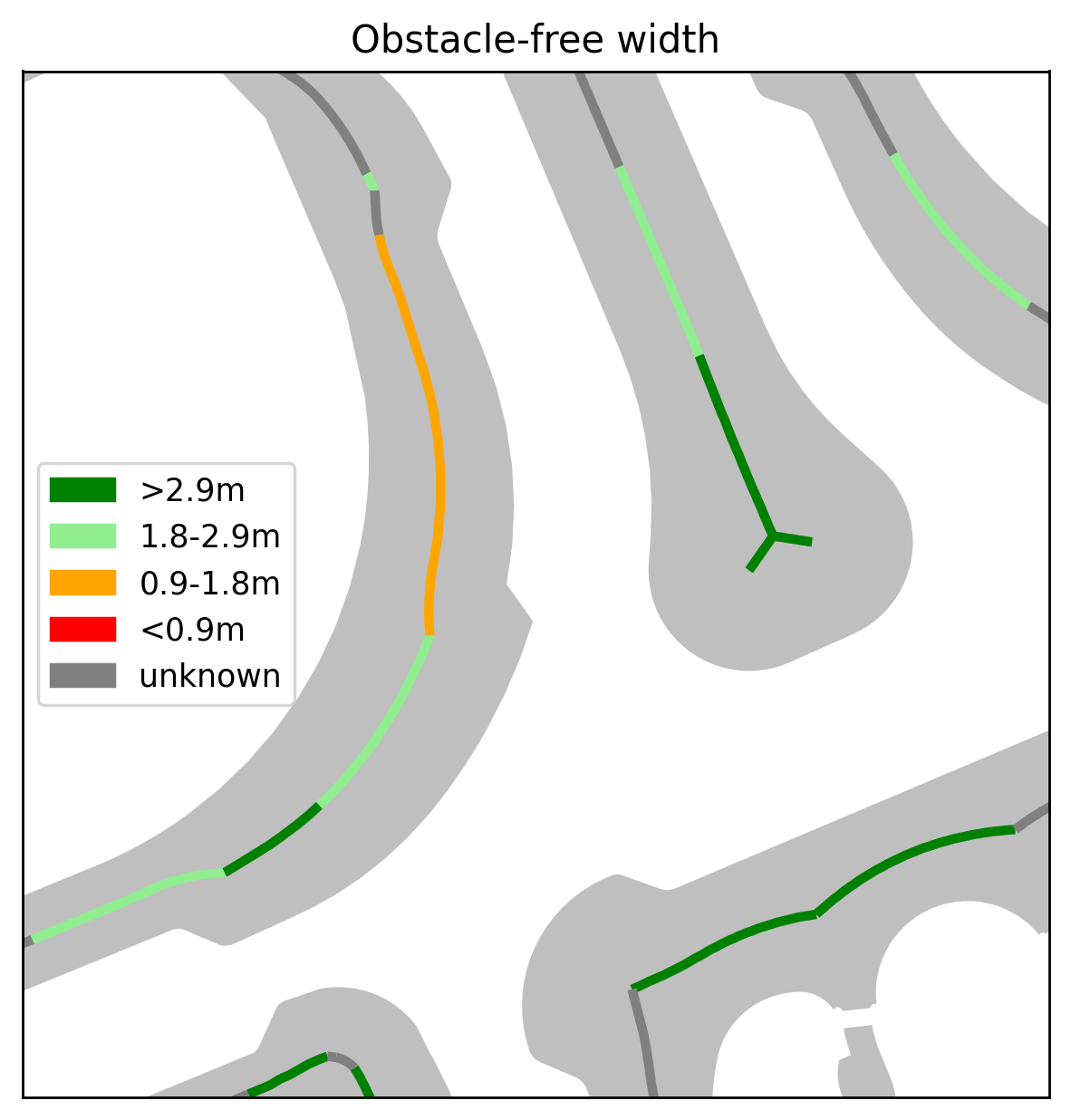} & \includegraphics[width=0.49\linewidth]{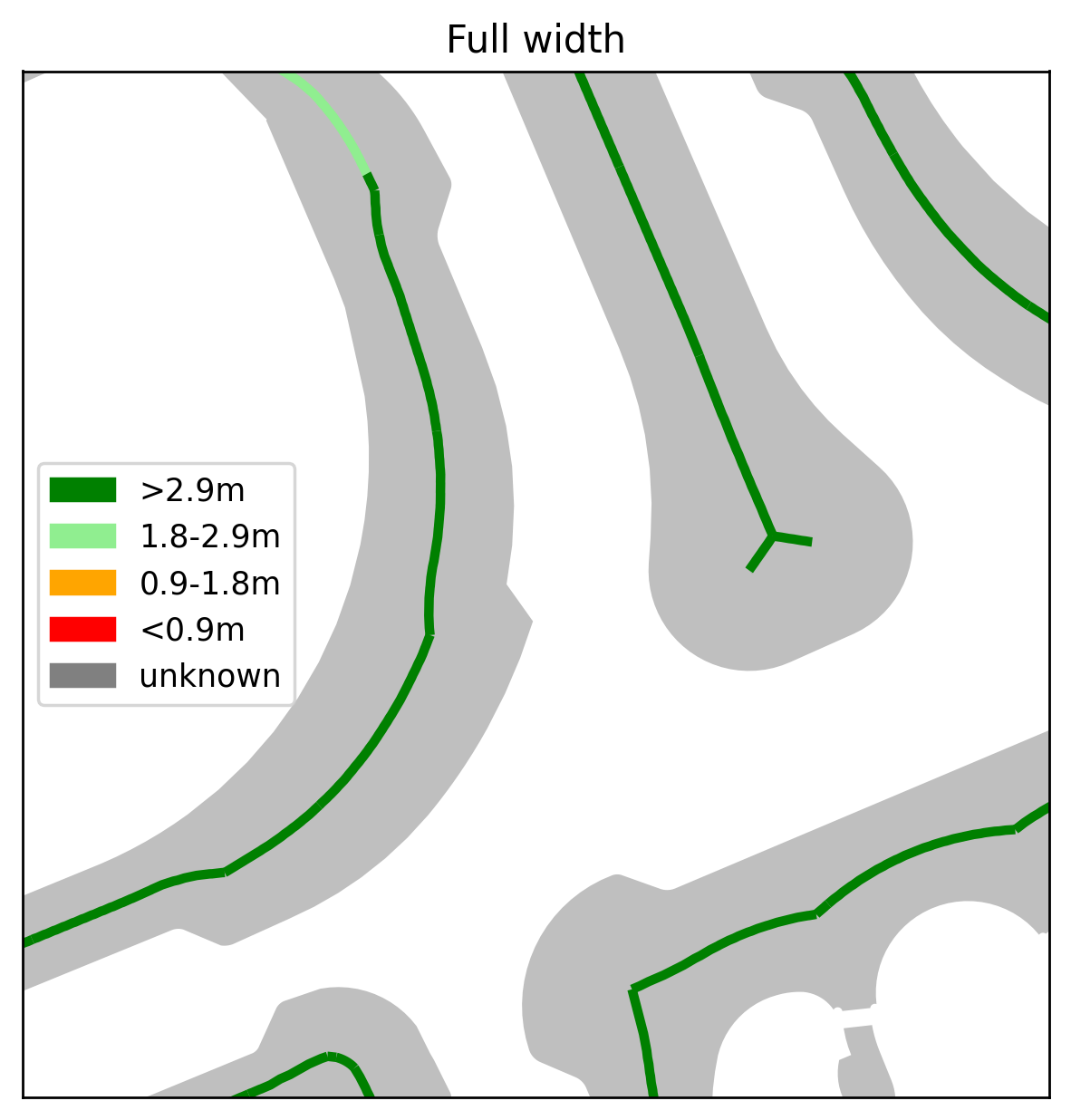} \\
  \end{tabular}
  \caption{Example map of output obstacle-free width and full widths of sidewalk segments (major paths).}
  \label{fig:finalmap}
\end{figure}

In order to compare the obstacle-free and full widths, we need a mapping from the detailed paths to the major paths.
To do so, we first create a graph based on all the detailed paths around obstacles. The distances on the edges of this graph are multiplied by a penalty term that is inversely exponentially proportional to the path widths (by category in the following bins $\{<0.9m, 0.9-1.8m, 1.8-2.9m, >2.9m\}$). Second, we obtain the start and end points of each major path. For these start and end points, we search for the nearest nodes in the graph. We then calculate an optimal route between the start and end node, using Dijkstra’s algorithm\footnote{As implemented in the NetworkX Python package} \cite{dijkstra1959note}, on a subset of the nodes in the graph that represent the sidewalk around the major path segment. Figure \ref{fig:2major} illustrates a major path segment in black, and in purple, the nodes of the optimal route found over minor paths.
For all major paths, we determine the obstacle-free width (category) as the minimal width of the edges over the optimal routes between their end points. As such we can compare the obstacle-free width and the full width for each major path. This results in an updated map as illustrated in Figure \ref{fig:finalmap}.
We note that there are major paths for which the width is unknown (grey), typically because the LIDAR scanner could not scan the corresponding sidewalk.

\section{Validation and Results}
\begin{figure}
    \centering
    \includegraphics[width=\linewidth]{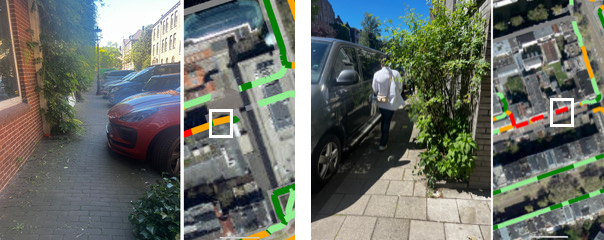}
    \caption{Example of validation. The left part of each panel shows the real situation, and the right part shows the computed sidewalk width with the white square showing the precise location.}
    \label{fig:validation}
    \vspace{-1em}
\end{figure}

To validate the outcomes, we first confirmed samples of the calculated widths digitally, by using Google Maps images and measuring distances in the point cloud data. Secondly, we did a physical validation. We visited several places in the city and compared the sidewalk width as experienced on the streets with our data (Figure \ref{fig:validation}). While the width categories that we computed were indeed accurate in the majority of cases, we do note that the data we used inherently provides a snapshot at a particular point in time. For example, we noticed that bicycles were placed differently at different moments, and plants and bushes had grown or were trimmed down. It is important to recognise these limitations when interpreting our results.

That being said, using our data we were able to compare our obstacle-free width estimates to the full widths as extracted from the municipal databases (BGT). We found that there are large discrepancies between the full widths and the obstacle-free widths. When we compare the percentage in the category for the full widths to the percentage in the corresponding obstacle-free width category (excluding unknown), we find that where $47.3\%$ of sidewalks have a known full width of over $2.9m$, this is only $26.6\%$ for the obstacle-free width. Furthermore, where only $17.7\%$ of the sidewalks have a full width of $0.9-1.8m$, this is no less than $32.8\%$ of the sidewalks for the obstacle-free width, and for $<0.9m$ this is $2.5\%$ versus $8.5\%$. Furthermore, there can be a large gap between the known full width and the estimated obstacle-free width. For example, $16\%$ of the sidewalks with an estimated obstacle-free width of $<0.9m$ had a full width of $>2.9m$, and $22\%$ a full width between $1.8$ and $2.9m$.

\section{Conclusion}
We have shown how we can estimate the accessible, i.e., obstacle-free, sidewalk width using point cloud data combined with topographical maps. Our results indicate that there can be large discrepancies between the full width and the obstacle-free widths of sidewalks. Therefore, we believe that obtaining information regarding the obstacle-free width is important, both for policy-making and for navigation for individuals. 

In future work, we aim to reduce the occurrence of ``unknown widths'' for paths where we lack data by using additional methods of LIDAR data gathering, e.g., by using scanners mounted on bicycles. We also want to deal more accurately with bridges, squares, entrances, parked bicycles, and the seasonality of plants. Another interesting direction for future work is combining the automated detection with crowdsourcing \cite{saha2019project} in order to improve and validate object detection. {\color{black} Finally, we hope to make cross-city and -country comparisons to validate our method.} 

We are currently working on turning these obstacle-free widths and full widths into a map, which we hope to publish soon. We are excited to add more features to this accessible sidewalk map in the future, such as curb heights, curb ramps, crosswalks, public transport stops and missing fences. Eventually, we hope this map can be used as a local route planner to assist people with reduced mobility. We note that this may require new multi-objective path planning algorithms, where different users can have different preferences with regards to the overall minimal widths, or the number of occurrences of small widths along their preferred route \cite{roijers2013survey,davoodi2013multi}.

\begin{acks}
We thank Frans Ost\'e (City of Amsterdam) and Willem van Waas (City of Amsterdam) for providing existing guidelines and policies related with pedestrian mobility and thinking along about the application of this data. We thank Jette Bolle (City of Amsterdam) and Eric Groot Kormelink (Clientenbelang) for their advice on accessibility in the local context. We also thank Dr Victor Pineda (World Enabled) for his advice on accessibility in a global context.
\end{acks}

\bibliographystyle{ACM-Reference-Format}
\bibliography{sidewalk}

\end{document}